\documentclass{article}
\usepackage{amsmath,graphicx,mlspconf}
\usepackage{amssymb}  
\usepackage{amsthm}   
\usepackage{algorithm}
\usepackage{algpseudocode}
\usepackage{xcolor}
\usepackage[pdfstartview=Fit]{hyperref}
\usepackage{bm}
\copyrightnotice{979-8-3503-2411-2/23/\$31.00 {\copyright}2025 IEEE}

 \toappear{2025 IEEE INTERNATIONAL WORKSHOP ON MACHINE LEARNING FOR SIGNAL PROCESSING, AUG. 31– SEP. 3, 2025, ISTANBUL, TURKEY}

\title{EMORF-II: Adaptive EM-Based Outlier-Robust Filtering With Correlated Measurement Noise}
\name{%
   Arslan Majal$^{\star}$%
   \qquad Aamir Hussain Chughtai$^{\star \star}$\
   \qquad Muhammad Tahir $^{\star \star \star}$\
}
\address{%
   $^{\star}$ University of Wisconsin Madison \
   $^{\star \star}$ University of Engineering and Technology, Lahore\\ %
   $^{\star \star \star}$ Lahore University of Management Sciences%
}

\begin{document}

\maketitle

\begin{abstract}
We present a learning-based outlier-robust filter for a general setup where the measurement noise can be correlated. Since it is an enhanced version of EM-based outlier robust filter (EMORF), we call it as EMORF-II. As it is equipped with an additional powerful feature to learn the outlier characteristics during inference along with outlier-detection, EMORF-II has improved outlier-mitigation capability. Numerical experiments confirm performance gains as compared to the state-of-the-art methods in terms of accuracy with an increased computational overhead. However, thankfully the computational complexity order remains at par with other practical methods making it a  useful choice for diverse applications.      
 \end{abstract}
\begin{keywords}
State-space models, Bayesian Inference, Outliers, Statistical Filtering, Expectation-Maximization.
\end{keywords}
\section{Introduction}

Bayesian filtering entails online estimation of the latent states of dynamical systems using noisy data. This challenge remains central to numerous disciplines e.g. robotics, sensor fusion, and target tracking etc. \cite{8517028,9596290,9563227}. Generally the nominal measurement noise is correlated in variety of applications including Real Time Kinematic (RTK) systems, sensors networks and time-difference-of-arrival (TDOA)-based systems \cite{li2020robust,10679915}. Assuming perfect apriori knowledge of the noise statistics, standard filters can handle these general scenarios \cite{sarkka2023bayesian}. 

In several real-world applications the data can be corrupted with unaccounted outliers crippling the standard filtering paradigm. This calls for outlier-robust filtering solutions. Traditionally, robustness is based on predefined assumptions about measurement noise statistics \cite{article1}, assuming fixed prior models for the residual errors between predicted and actual sensor data \cite{article2}, and applying thresholding techniques \cite{article3}. However, such techniques necessitate user tuning, and the performance of these algorithms becomes sensitive to the chosen parameters \cite{nakabayashi2019nonlinear}. This challenge motivates the design of learning-based, tuning-free methods that assume a prior structure of the measurement model while dynamically learning its parameters. The state along with parameters describing the data (including outliers) are estimated jointly thereby reducing the dependence on user set parameters \cite{10679915,9716131}.

Due to the promise offered by learning-based outlier-robust filters, this approach of devising Bayesian filters has been actively adopted recently. Variational Bayesian (VB) techniques are typically favored over Particle Filters (PFs) in their designs as they are less computationally intensive and allow reuse of standard Gaussian filtering results. In this regard, we proposed the Selective Outlier Rejection (SOR) filter \cite{9716131} that utilizes a vector-parameterized measurement covariance matrix to mitigate the impact of outliers in sensor measurements. While this approach has demonstrated superior performance compared to methods based on scalar-parameterized covariance matrices \cite{6349794}, it assumes that the noise in the measurement vector is uncorrelated an assumption that may be violated in several real-world scenarios \cite{article6,inproceedings1}. To account for the effects of correlated noise in the measurements, we proposed another VB-based filter namely Expectation-Maximization (EM)-based outlier robust filter (EMORF) \cite{10679915}. This method assumes a prior model for measurements that accommodates correlated noise while simultaneously incorporating a vectorized outlier detection scheme. Due to its model, EMORF has shown better results as compared to a similar method Variational Bayes
Kalman Filters (VBKFs) \cite{li2020robust} and is simpler to implement.

Built on a model inspired from the SOR filter, EMORF estimates the state and parameters to detect the outlier during inference. However, it does not learn the outlier characteristics. This information can be useful and if utilized properly can result in better estimators as demonstrated in our Adaptive Selective Outlier Rejecting (ASOR) method \cite{10430181}. With this motivation we present EMORF-II, which not only handles correlated measurement noise and detects outliers but also learns the the characteristics of outliers. To this end, we take inspiration from the ASOR method to construct EMORF-II. We demonstrate the performance gains of EMORF-II compared with similar state-of-the-art outlier-robust learning-based methods that address correlated measurement noise through numerical experiments. We use the same notations as in EMORF. For completeness, we present the notations, details of derivations and code in the 
\href{https://github.com/chughtaiah/EMORF-II}{supplementary material}.
\\

\section{STATE-SPACE MODEL}
\begin{figure}[h!] 
    \centering
\includegraphics[width=0.4\textwidth, trim=15pt 5pt 5pt 20pt, clip]{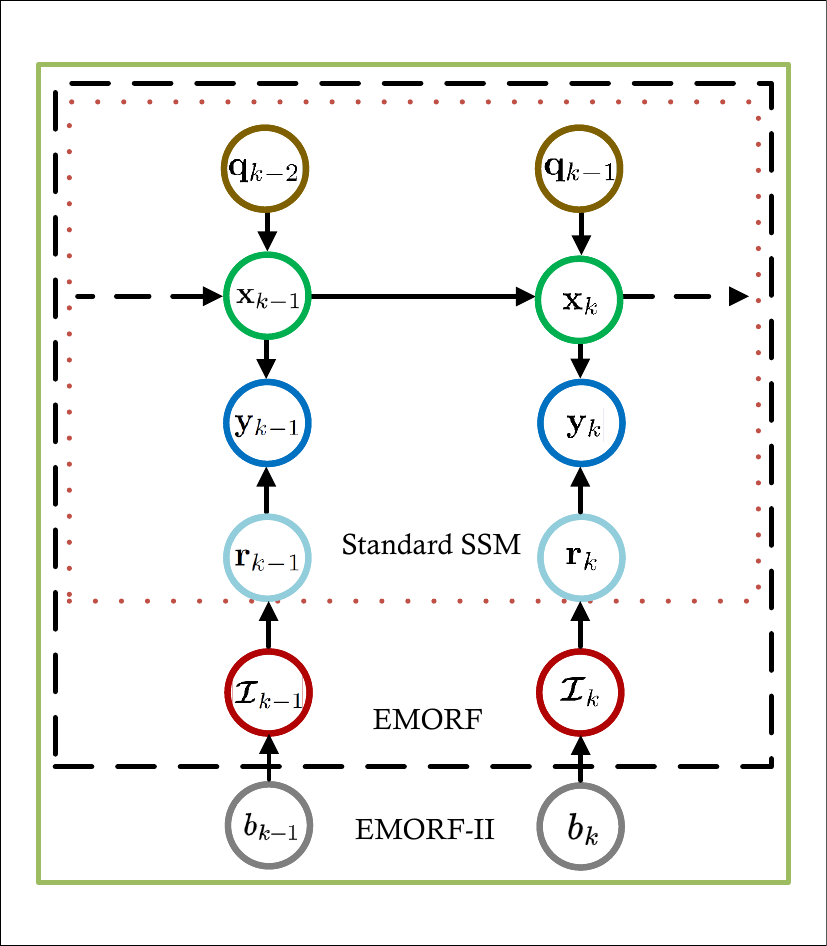} 
    \caption{Probabilistic graphical model for EMORF-II}
    \label{PGM}
\end{figure}
To devise EMORF-II, we modify the state-space model (SSM) model of EMORF as shown in the graphical model in Fig.~\ref{PGM}. Our modification is inspired from the results of the Adaptive Selective Outlier-Rejecting (ASOR) method which introduces a hierarchical model to describe the system dynamics and characteristics of outliers \cite{10430181}. The model results in a state estimator that is able to \textit{learn} the covariance of outliers along with their occurrence.  

\subsection{Proposed Model}
The proposed SSM for EMORF-II is given as
\begin{align}  \bm{x}_{k} & =\bm{f}(\bm{x}_{k-1})+\bm{q}_{k-1} \label{1} \\ \bm{y}_{k} & =\bm{h}(\bm{x}_{k})+\bm{r}_{k} \label{2} \end{align}
where $\bm{x}_{k} \in \mathbb{R}^n$ is the state vector, $\bm{y}_{k} \in \mathbb{R}^m$ is the measurement vector, $\bm{f}(\cdot): \mathbb{R}^n \rightarrow \mathbb{R}^n$ is the process model and $\bm{h}(\cdot): \mathbb{R}^n \rightarrow \mathbb{R}^m$ represents the measurement model. Furthermore, the process noise is normally distributed, as in the standard SSM, such that $\bm{q}_k \sim \mathcal{N}(\bm{q}_k\big| \bm{0}, \bm{Q}_k)$. 

Similar to EMORF, the measurement noise is supposed to follow a Gaussian sum distribution to capture the outlier effects such that $\bm{r}_{k}\sim \sum_{\boldsymbol{\mathcal{I}}_k}  \textit{p}\left(\bm{r}_k\big| \boldsymbol{\mathcal{I}}_k\right)\textit{p}\left(\boldsymbol{\mathcal{I}}_k\right)$ with
\begin{align} \textit{p}\left(\bm{r}_k\big| \boldsymbol{\mathcal{I}}_k\right)=\mathcal{N}\left(\bm{r}_k \big|  \bm{0}, \bm{R}_k\left(\boldsymbol{\mathcal{I}}_k\right)\right) \label{rkIk} \end{align}
with $\delta(.)$ denoting the delta function, the measurement covariance matrix $ \bm{R}_k(\boldsymbol{\mathcal{I}}_k)$ is given as 
{\footnotesize
\[
\begin{bmatrix}
\dfrac{R_k^{1,1}}{\mathcal{I}_k^1} 
  & \cdots 
  & R_k^{1,m}\,\delta(\mathcal{I}_k^1-1)\,\delta(\mathcal{I}_k^m-1) \\
\vdots & \ddots & \vdots \\
R_k^{m,1}\,\delta(\mathcal{I}_k^m-1)\,\delta(\mathcal{I}_k^1-1) 
  & \cdots 
  & \dfrac{R_k^{m,m}}{\mathcal{I}_k^m}
\end{bmatrix}
\]
}

 As in EMORF, we model the presence (or absence) of outliers through an indicator random vector $\boldsymbol{\mathcal{I}}_k \in \mathbb{R}^m$. $\mathcal{I}_k^i \neq 1$ represents the presence of an outlier in the $i$-th dimension of measurement at time step $k$ whereas $\mathcal{I}_k^i = 1$ represents the absence of an outlier. For inferential tractability and due to lack of prior knowledge of correlations between outliers, we assume that outliers occur independently in each dimension as in EMORF. Similar to the ASOR method, we hierarchically model $\boldsymbol{\mathcal{I}}_k$ as
\begin{align}
  &p\bigl(\boldsymbol{\mathcal{I}}_k \big|  b_k\bigr)=\prod_{i=1}^m p\bigl(\mathcal{I}_k^i \big|  b_k\bigr) \label{modelI}
\end{align}
with
\begin{align}
  p\bigl(\mathcal{I}_k^i\big|  b_k\bigr) =
       (1-\theta_k^i)\overbrace
       {f(a_k,b_k)(\mathcal{I}_k^i)^{a_k-1}e^{-b_k\mathcal{I}_k^i}}^{Gamma(a_k,b_k)}
       +\delta\bigl(\mathcal{I}_k^i-1\bigr)\theta_k^i \label{pikbk} \nonumber
\end{align}
where the two components of $\textit{p}\left(\mathcal{I}_k^i \big|  b_k\right)$ are assumed disjoint with the Gamma density defined as zero for
$\mathcal{I}_k^i=1$. $\theta_k^i$ is the prior probability of no outlier occurring in the $i$-th dimension. To model outliers, $\mathcal{I}_k^i$ is supposed to obey a Gamma distribution supported on the set of positive real numbers i.e. $\mathcal{I}_k^i \sim \text{Gamma}(a_k, b_k)$ with prior parameters $a_k,b_k$ and $f(a_k, b_k) = ({b_k^{a_k}}/{\Gamma(a_k)})$. The  parameter $b_k$ of the Gamma distribution captures the characteristics of outliers. Since the conjugate
prior of $b_k$ is also a Gamma distribution, we use this as the prior on $b_k$ with parameters $A_k$  and $B_k$ given as
\begin{equation}
\label{b_prior}
    \textit{p}(b_k) = f(A_k, B_k)  b_k^{A_k-1} e^{-B_k b_k}
\end{equation}

\section{OUTLIER-ROBUST FILTERING}

In Bayesian filtering, the goal is to compute $ \textit{p}(\bm{x}_{k} \big|  \bm{y}_{1:k})$ i.e. the posterior distribution of the state given the set of observations. Using Bayes' rule for our modeling setup, the joint posterior $\textit{p}(\bm{x}_{k}, \boldsymbol{\mathcal{I}}_k, b_k \big|  \bm{y}_{1:k})$ can be written as 
\begin{equation}
\textit{p}(\bm{x}_{k}, \boldsymbol{\mathcal{I}}_k, b_k \big|  \bm{y}_{1:k}) \propto \textit{p}(\bm{y}_{k} \big|  \boldsymbol{\mathcal{I}}_k, x_k) \textit{p}(\bm{x}_{k}\big| \bm{y}_{1:k-1}) \textit{p}(\boldsymbol{\mathcal{I}}_k\big| b_k) \textit{p}(b_k) \label{joint}
\end{equation}

Since marginalizing of the joint posterior to obtain the posterior of the state is intractable, we employ the Expectation Maximization (EM) to this end as \cite{murphy2012machine}

\subsection{EM for inference}
At each time instant, the E and M steps are recursively invoked till convergence to approximate the state posterior. In particular, in the E-step we approximate the state distribution as   

\textbf{E-Step}
\begin{equation}
\textit{q}(\bm{x}_{{k}})=\textit{p}(\bm{x}_{k} \big|  \bm{y}_{1: k}, \hat{\boldsymbol{\mathcal{I}}}_k,\hat{b}_k) \propto \textit{p}(\bm{x}_{k}, \hat{\boldsymbol{\mathcal{I}}}_k,\hat{b}_k \big|  \bm{y}_{1: k}) \label{Estep}
\end{equation}

The estimates of other parameters $\hat{b}_k$ and $\hat{\mathcal{I}}_k^i \ \forall \ i$ are obtained in the M-steps as

\textbf{M-Steps}
\begin{equation}
\hat{\mathcal{I}}_k^i=\underset{\mathcal{I}_k^i}{\operatorname{argmax}}\big\langle\ln (\textit{p}(\mathrm{x}_k, \mathcal{I}_k^i, \hat{\boldsymbol{\mathcal{I}}}_k^{i-},\hat{b}_k \big|  \bm{y}_{1: k})\big \rangle_{\textit{q}(\bm{x}_{k})}
\label{Mstep2}
\end{equation}

\begin{equation}
\hat{b}_k=\underset{{b}_k^i}{\operatorname{argmax}}\big\langle\ln (\textit{p}(\mathrm{x}_k, \hat{\boldsymbol{\mathcal{I}}}_k,{b}_k \big|  \bm{y}_{1: k})\big \rangle_{\textit{q}(\bm{x}_{k})} \label{Mstep1}
\end{equation}
where $\hat{\boldsymbol{\mathcal{I}}}_k^{i-}$ contains the all the elements of the estimated vector $\hat{\mathcal{I}}_k$ except the $i$-th element.

\subsection{State Estimation}
To invoke the E-step in \eqref{Estep} for approximating the posterior distribution of $\bm{x}_{k}$, we need to evaluate the joint posterior in \eqref{joint}. This requires approximating the predictive distribution $\textit{p}(\bm{x}_{k} \big|  \bm{y}_{1:k-1})$ at each time step. This is achieved through the prediction step described as follows.

\textbf{Prediction:}
Using the process model and the estimated state posterior distribution from the previous time step 
$\textit{p}(\bm{x}_{k-1} \big|  \bm{y}_{1:k-1})$ we can estimate the predictive density. To leverage the standard filtering results, we assume the predictive distribution is Gaussian given as \cite{sarkka2023bayesian}
\begin{equation}
\label{x_prior}
 \textit{p}(\bm{x}_{k} \big|  \bm{y}_{1:k-1}) \approx \mathcal{N}(\bm{x}_{k} \big|  \bm{m}_k^-, \bm{P}_k^-)    
\end{equation}
where the parameters of the predictive distribution are
\begin{align}
\label{m_pred}
\bm{m}_k^- = & \int f(\bm{x}_{k-1}) \mathcal{N}(x_{k-1} \big|  \bm{m}_{k-1}^+, \bm{P}_{k-1}^+) \, d\bm{x}_{k-1}\\
\label{p_pred}
\bm{P}_k^- = & \int (f(\bm{x}_{k-1}) - \bm{m}_k^-)(f(\bm{x}_{k-1}) - \bm{m}_k^-)^\top \nonumber
\\
&\ \ \ \mathcal{N}(\bm{x}_{k-1} \big|  \bm{m}_{k-1}^+, \bm{P}_{k-1}^+) \, d\bm{x}_{k-1} + \bm{Q}_{k-1} 
\end{align}


\textbf{Update:}
The joint distribution in \eqref{joint} can be expressed as 
{\small
\begin{align}
\label{post_ful}
&\textit{p}(\bm{x}_{k}, \boldsymbol{\mathcal{I}}_k, b_k \big|  \bm{y}_{1:k}) \propto \frac{\mathcal{N}(\bm{x}_{k} \big|  \bm{m}_k^-, P_k^-)}{\sqrt{(2\pi)^m \big| R_k(\boldsymbol{\mathcal{I}}_k)\big| }} \exp\big( \text{-} \frac{1}{2} (\bm{y}_{k} - h(\bm{x}_{k}))^\top  \nonumber \\
& R_k^{-1}(\boldsymbol{\mathcal{I}}_k) (\bm{y}_{k} - h(\bm{x}_{k})) \big) \prod_i \left[(1-\theta_k)f(a_k, b_k) (\mathcal{I}_k^i)^{a_k-1} e^{-b_k\mathcal{I}_k^i} + \right. \nonumber \\
&\left. \theta_k^i \delta(\mathcal{I}_k^i - 1) \right] f(A_k, B_k)  b_k^{A_k-1} e^{-B_k b_k}
\end{align}
}
where we have used the expressions for the prior distributions from \eqref{modelI}, \eqref{b_prior} and the conditional measurement likelihood  considering \eqref{2} and \eqref{rkIk}.
To approximate the posterior distribution for $\bm{x}_{k}$ we use \eqref{Estep} and \eqref{post_ful} to arrive at
\begin{align}
&\textit{q}(\bm{x}_{k}) \propto \exp\big( -\tfrac{1}{2} (\bm{y}_{k} - h(\bm{x}_{k}))^\top \bm{R}_k^{-1}(\hat{\boldsymbol{\mathcal{I}}}_k) (\bm{y}_{k} - h(\bm{x}_{k}))  \nonumber \\
& -\tfrac{1}{2} (\bm{x}_{k} - \bm{m}_k^-)^\top (\bm{P}_k^-)^{-1} (\bm{x}_{k} - \bm{m}_k^-) \big)
\end{align}
where $\bm{R}_k^{-1}(\hat{\boldsymbol{\mathcal{I}}}_k)$ is the matrix obtained by inverting $\bm{R}({\boldsymbol{ \hat{\mathcal{I}}} }_k)$ after plugging in the EM estimates $\boldsymbol{\hat{\mathcal{I}}}_k$  from VB iterations. Finally to approximate $\textit{q}(\bm{x}_{k})$ as a normal distribution \( \mathcal{N}(\bm{x}_{k} \big|  \bm{m}_k^+, \bm{P}_k^+) \) we employ the general Gaussian filtering results which provide \cite{sarkka2023bayesian}
\begin{align}
\label{m_est}
    & \bm{m}_k^+ = \bm{m}_k^- + \bm{K}_k (\bm{y}_{k} - \mu_k)\\
    \label{P_est}
    & \bm{P}_k^+ = \bm{P}_k^- - \bm{C}_k \bm{K}_k^\top
\end{align}
where
\begin{align}
    & \bm{K}_k = \bm{C}_k (\bm{U}_k + \bm{R}_k(\hat{\boldsymbol{\mathcal{I}}}_k))^{-1}\\
    & \boldsymbol{\mu}_k = \int h(\bm{x}_{k}) \mathcal{N}(\bm{x}_{k} \big|  \bm{m}_k^-, \bm{P}_k^-) \, d\bm{x}_{k}\\
    & \bm{U}_k = \int (h(\bm{x}_{k}) - \boldsymbol{\mu}_k)(h(\bm{x}_{k}) - \boldsymbol{\mu}_k)^\top \mathcal{N}(\bm{x}_{k} \big|  \bm{m}_k^-, \bm{P}_k^-) \, d\bm{x}_{k}\\
    & \bm{C}_k = \int (\bm{x}_{k} - \bm{m}_k^-)(h(\bm{x}_{k}) - \boldsymbol{\mu}_k)^\top \mathcal{N}(\bm{m}_k \big|  \bm{m}_k^-, \bm{P}_k^-) \, d\bm{x}_{k}
\end{align}

\subsection{Parameter Estimation}
We can obtain the estimates of all the elements of $\boldsymbol{\hat{\mathcal{I}}}_k$ given as $\hat{\mathcal{I}}^{i}_{k}$ successively, using the M-step in \eqref{Mstep2} which leads to

{\small
\begin{align}
&\hat{\mathcal{I}}^{i}_{k} = \operatorname*{argmax}_{\mathcal{I}^{i}_{k}} \Bigl\{ -\frac{1}{2} \operatorname{tr}\bigl(\bm{W}_{k}\bm{R}^{-1}_{k}(\mathcal{I}^{i}_{k},\hat{\boldsymbol{\mathcal{I}}}^{i-}_{k})\bigr) - \frac{1}{2}\ln\big| \bm{R}_{k}(\mathcal{I}^{i}_{k},\hat{\boldsymbol{\mathcal{I}}}^{i-}_{k})\big|  \nonumber \\
&+ \ln\bigl((1-\theta_{k})f(a_k, \hat{b}_k) (\mathcal{I}_k^i)^{a_k-1} e^{-\hat{b}_k\mathcal{I}_k^i} + \theta_{k}\delta(\mathcal{I}^{i}_{k}-1)\bigr) + k_1 \Bigr\}  \label{IKi1}
\end{align} 
}where we use the property of the trace operator applied on
the product of matrices given as $\operatorname{tr}(\bm{ABC}) = \operatorname{tr}(\bm{CAB})$ and $k_1$ is a constant. $\bm{R}_{k}(\mathcal{I}^{i}_{k},\hat{\boldsymbol{\mathcal{I}}}^{i-}_{k})$ denotes $\bm{R}_{k}(\boldsymbol{\mathcal{I}}_{k})$ evaluated at $\boldsymbol{\mathcal{I}}_{k}$ with its $i$-th element as $\mathcal{I}^{i}_{k}$ and remaining entries $\hat{\boldsymbol{\mathcal{I}}}^{i-}_{k}$ and $\bm{W}_{k}$ is given as
\begin{equation}
\label{W_est}
\bm{W}_{k} = \int (\bm{y}_{k} - \bm{h}(\bm{x}_{k}))(\bm{y}_{k} - \bm{h}(\bm{x}_{k}))^{\top} \mathcal{N}(\bm{x}_{k}\big| \bm{m}^{+}_{k},\bm{P}^{+}_{k}) d\bm{x}_{k}
\end{equation}

By maximizing the objective function in \eqref{IKi1}, we end up with the following decision criterion for all the $\hat{\mathcal{I}}^{i}_{k}$
\begin{align}
\label{Ik_est}
& \mathcal{I}_k^i = 
\begin{cases} 
1 & \text{if } H_k^i/G_k^i \geq 1 \\ 
({\alpha_k -1})/{\beta_k^i} & \text{if } H_k^i/G_k^i < 1 
\end{cases}
\end{align}
where $H_k^i$ and $G_k^i$ are given as 

{\small
\begin{align}
\label{Hk}
H_k^i = &\exp\Bigl(
  -\tfrac{1}{2}\ln\bigl\lvert \bm{R}_{k}(\mathcal{I}_{k}^{i}=1,\hat{\boldsymbol{\mathcal{I}}}_k^{-i})\bigr\rvert\\
  \nonumber
  & \quad -\tfrac{1}{2}\,\mathrm{tr}\bigl(\bm{W}_{k}\,\bm{R}_{k}^{-1}(\mathcal{I}_{k}^{i}=1,\hat{\boldsymbol{\mathcal{I}}}_k^{-i})\bigr)
\Bigr)\,\theta_{k}^{i}\\
\label{Gk}
G_k^i = &
  \bigl(R_{k}^{ii}\bigr)^{-\tfrac{1}{2}}
  \bigl\lvert\hat{\bm{R}}_{k}^{-i,-i}\bigr\rvert^{-\tfrac{1}{2}}
  \exp\Bigl(
    -\tfrac{1}{2}\,\mathrm{tr}\bigl(\bm{W}_{k}^{-i,-i}\,(\hat{\bm{R}}_{k}^{-i,-i})^{-1}\bigr)
  \Bigr)
  \nonumber\\
&\quad
  (1-\theta_{k}^{i})
  \frac{\Gamma(\alpha_{k})\,\hat{b}_k^{\,a_k}}
       {\Gamma(a_{k})\,(\beta_k^i)^{\alpha_k}}
\end{align}
}
where for any matrix $\bm{R}$ we obtain the sub-matrix  $\bm{R}^{-i, -i}$ by removing its $i$-th column and row. $\alpha_k$ and $\beta_k^i$ are given as 
\begin{align}
&\alpha_k = a_k + 0.5
\\
&\beta_k^i = \hat{b}_k + 0.5{W_{k}^{ii}}/{{R_{k}^{ii}}}
\end{align}
Using the M-step in \eqref{Mstep1}, we can arrive at the following expression for estimating $\hat{b}_k$ as 
\begin{align}
\label{b_est}
\hat{b}_k = ({\overline{A}_k - 1})/{\overline{B}_k}
\end{align}
where 
\[
\overline{A}_k = M_k a_k + A_k \quad \text{and} \quad \overline{B}_k = B_k + \sum_{\{i : \hat{\mathcal{I}}_k^i \neq 1\}} \hat{\mathcal{I}}_k^i
\]

In this formulation, the summation defining \(\overline B_k\) includes only those indices \(i\) for which \(\hat{\mathcal{I}}_k^i\neq1\), thereby excluding any terms with \(\hat{\mathcal{I}}_k^i=1\).  Further, \(M_k=\#\{\,i:\hat{\mathcal{I}}_k^i\neq1\}\) or the count of $\boldsymbol{\mathcal{I}_k}$ elements not equal to one. 

The pseudocode for the resulting algorithm EMORF-II is shown in Algorithm \ref{alg:asor}. For convergence, we check if the normalized $l_2$ norm of changes in state estimates during EM iterations falls below a predefined threshold as proposed in EMORF.


\begin{algorithm}
  \caption{EMORF-II}
  \label{alg:asor}
  \begin{algorithmic}[1]
  \State Initialize $\bm{m}_{0}^{+}\, , \bm{P}_{0}^{+}$
    \For{$k = 1,\dots,K$}
\State Initialize {\footnotesize  $A_k,B_k,a_k,\hat b_k,\bm Q_k,\bm R_k\;\forall\ k$, $\hat{\mathcal I}_k^i=1\;\forall\ i$, 
 $\theta_k^i\;\forall\ i$}
      \State Evaluate $\alpha_k = a_k + 0.5$
      \State \textbf{Prediction}
       \State Evaluate $\bm{m}_k^{-}$ and $\bm{P}_k^{-}$ using \eqref{m_pred} and \eqref{p_pred}
       \State \textbf{Update}
      \While{the convergence criterion has not been met}
        \State Update $\bm{m}_k^{+}$ and $\bm{P}_k^{+}$ using \eqref{m_est} and \eqref{P_est}
        \State Update $\hat b_k$ using \eqref{b_est}
        \State Update $\hat{\mathcal{I}_k^i}\ \forall\,i$ using \eqref{Ik_est}
      \EndWhile
    \EndFor
  \end{algorithmic}
\end{algorithm}

\section{Numerical Experiments}
\subsection{Evaluation Setup}
To demonstrate the capabilities of EMORF-II, we conduct a series of numerical experiments and compare the results with those of recent state-of-the-art VB-based outlier-robust filters.  All experiments are performed using \textsc{Matlab}~R2023b on a Windows 11 laptop equipped with an \mbox{Intel i9-13900H} processor and 32{GB} of RAM.

We assume that a target moves according to the following nonlinear state equation proposed in~\cite{8398426}
{\footnotesize
\begin{equation}
  \bm{x}_{k}=  
  \begin{bmatrix}
    1 & \dfrac{\sin(\omega_{k-1}\zeta)}{\omega_{k-1}} & 0 & \dfrac{\cos(\omega_{k-1}\zeta)-1}{\omega_{k-1}} & 0 \\
    0 & \cos(\omega_{k-1}\zeta)                       & 0 & -\sin(\omega_{k-1}\zeta)                      & 0 \\
    0 & \dfrac{1-\cos(\omega_{k-1}\zeta)}{\omega_{k-1}} & 1 & \dfrac{\sin(\omega_{k-1}\zeta)}{\omega_{k-1}} & 0 \\
    0 & \sin(\omega_{k-1}\zeta)                        & 0 & \cos(\omega_{k-1}\zeta)                       & 0 \\
    0 & 0                                             & 0 & 0                                             & 1
  \end{bmatrix}
  \bm{x}_{k-1} + \bm{q}_{k-1}
\end{equation}}
where  \(\zeta\) is the sampling period and the state vector
\(
  \bm{x}_{k}
  =
  [x_{k},\,\dot x_{k},\,y_{k},\,\dot y_{k},\,\omega_{k}]^{\mathsf T}
\)
contains planar position, velocity, and turn rate respectively.  The process noise is Gaussian,
\(
  \bm{q}_{k-1}\!\sim\!\mathcal{N}(\bm 0,\bm Q_{k-1}),
\)
with covariance given as \cite{8398426} 
\[
  \bm Q_{k-1}
  =
  \begin{bmatrix}
    \eta_{1}\bm M & \bm 0 & \bm 0\\
    \bm 0 & \eta_{1}\bm M & \bm 0\\
    \bm 0 & \bm 0 & \eta_{2}
  \end{bmatrix},
  \qquad
  \bm M
  =
  \begin{bmatrix}
    \zeta^{3}/3 & \zeta^{2}/2\\
    \zeta^{2}/2 & \zeta
  \end{bmatrix}
\]


Similar to EMORF, Time-Difference-Of-Arrival (TDOA) readings are used to estimate the state of the target. To that end, a network of \(m\) Time Of Arrival (TOA)-based range sensors laid out in a zig-zag pattern is assumed for measurements.  The \(i\)-th sensor is placed at
\(
  (\xi^{\rho_i},\eta^{\rho_i})
  =
  \bigl(350(i-1),\,350\!\bigl((i-1)\bmod2\bigr)\bigr)
\)
for \(i=1,\dots,m\). Designating sensor~1 as reference produces \(m-1\) TDOA ranges. To emulate corrupted data, we add the an outlier vector \(\bm o_{k}\) in the nominal noise as
\begin{align}
\label{y_out}
&  \bm y_{k} = \bm h(\bm x_{k})+\bm r_{k}+\bm o_{k}
\end{align}
where
\begin{align}
\nonumber
 & h^{j}(\bm x_{k})
  =
  \sqrt{(x_{k}-\xi^{\rho_{1}})^{2}+(y_{k}-\eta^{\rho_{1}})^{2}}
  \;-\;
  \\
  \label{h_x}
 & \sqrt{(x_{k}-\xi^{\rho_{j+1}})^{2}+(y_{k}-\eta^{\rho_{j+1}})^{2}}
  \quad j=1,\dots,m-1.
\end{align}

Consequently, the nominal covariance of \(\bm r_{k}\) is fully populated
\[
  \bm R_{k}
  =
  \begin{bmatrix}
    \sigma_{1}^{2}+\sigma_{2}^{2} & \cdots & \sigma_{1}^{2}\\
    \vdots & \ddots & \vdots\\
    \sigma_{1}^{2} & \cdots & \sigma_{1}^{2}+\sigma_{m}^{2}
  \end{bmatrix}
\]
where \(\sigma_{i}^{2}\) denotes the noise variance of sensor~\(i\). Moreover, we assume that \(\bm o_{k}\) follows the distribution
\begin{equation}
  \label{pok}
  \textit{p}(\bm o_{k})
  =
  \prod_{j=1}^{m-1}
  \mathcal J_{k}^{j}\,
\mathcal{N}\!\bigl(o_{k}^{j}\,\big| \,0,\gamma(\sigma_{1}^{2}+\sigma_{j}^{2})\bigr) 
\end{equation}
where \(\mathcal J_{k}^{j}\in\{0,1\}\) is a Bernoulli indicator marking an outlier in the \(j\)-th channel. Let \(\lambda\) be the probability that a single TOA measurement is corrupted. Since every TDOA involves the first TOA measurement as reference, the chance of observing no outlier in every TDOA dimension is \((1-\lambda)^{2}\). The factor \(\gamma\) inflates the nominal variance to produce the effect of outliers.  


\subsection{Comparative Methods and Parameters}
To evaluate the performance, we compare our proposed EMORF-II, against several baseline approaches. First, we consider a hypothetical outlier-robust Gaussian filter assuming perfect a priori knowledge of all outlier instances. Furthermore, we assess two variations of the generalized VBKF, namely, Gen. VBKF with $N = 10$ and Gen. VBKF with $N = 1$ ($N$ is a hyperparameter of the VBKF filter) together with the independent VBKF estimator (Ind. VBKF) from \cite{li2020robust}. We also consider EMORF for comparison.


For simulations, we assume $\bm{x}_{0}=[0,1,0,-1,-0.0524]^{\mathrm{T}}$, $\zeta=1$, $\eta_{1}=0.1$, $\eta_{2}=1.75\times 10^{-4}$, $\sigma_{j}^{2}=10$ and $K = 100$. All estimators are initialized with $\bm{m}_{0}^{+}\sim\mathcal{N}(\bm{x}_{0},\bm{P}_{0}^{+})$, $\bm{P}_{0}^{+}=\bm{Q}_{k}$ and use the Unscented Kalman Filter (UKF) \cite{wan2001unscented} as the core Gaussian filter. UKF parameters are set as $\alpha_{UKF}=1$, $\beta_{UKF}=2$, $\kappa_{UKF}=0$. We use a convergence threshold of $10^{-4}$ and set $\theta_k^i=0.5\ \forall\ i,k$ as neutral prior for absence of outliers as in EMORF. Other initialization parameters of EMORF-II are set as $A_k=10000, B_k=1000, a_k=1, \hat{b}_k=10000$ as proposed in ASOR. The numerical experiments are performed for 100 Monte Carlo (MC) runs to capture the error statistics in each scenario.

\subsection{Evaluation }
\begin{figure}[h!]
    \centering
\includegraphics[width=0.48\textwidth]{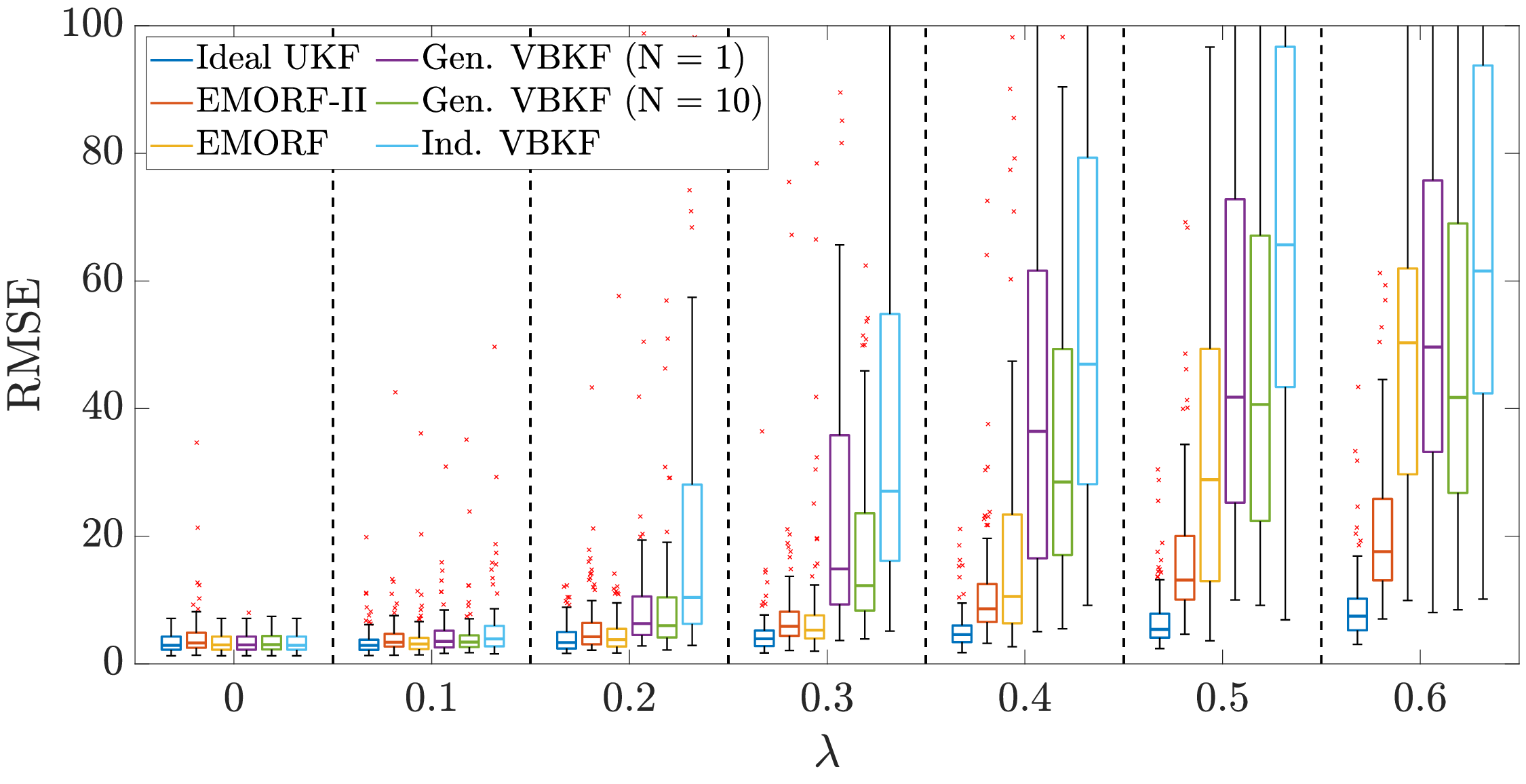} 
    \caption{RMSE comparison for $m = 5, \gamma = 1000,$ and $\lambda \in \{ 0, 0.1 ,\hdots 0.6\} $}
    \label{sim_1}
\end{figure}
First, we evaluate the accuracy of all the considered algorithms with increasing outlier occurrence probabilities denoted by $\lambda$. We set the parameters $m = 5$ and $\gamma = 1000$ in \eqref{pok}. The resulting error distributions considering the MC runs are visualized using box plots in Fig.~\ref{sim_1}. As expected, the ideal hypothetical UKF, having perfect knowledge of outlier occurrences, achieves the best performance. Notably, EMORF-II consistently results in lower Root-Mean-Squared-Error (RMSE) as compared to other practical methods as $\lambda$ increases. This improvement stems from its ability to adaptively learn outlier characteristics and use this information for their mitigation. In contrast, EMORF exhibits progressively larger estimation errors as outliers become more frequent, highlighting its limitation to dynamically learn the outlier characteristics. The performance of other estimators get even worse with increasing $\lambda$ (see experimental section of EMORF for more details).

\begin{figure}[h!]
    \centering
\includegraphics[width=0.48\textwidth]{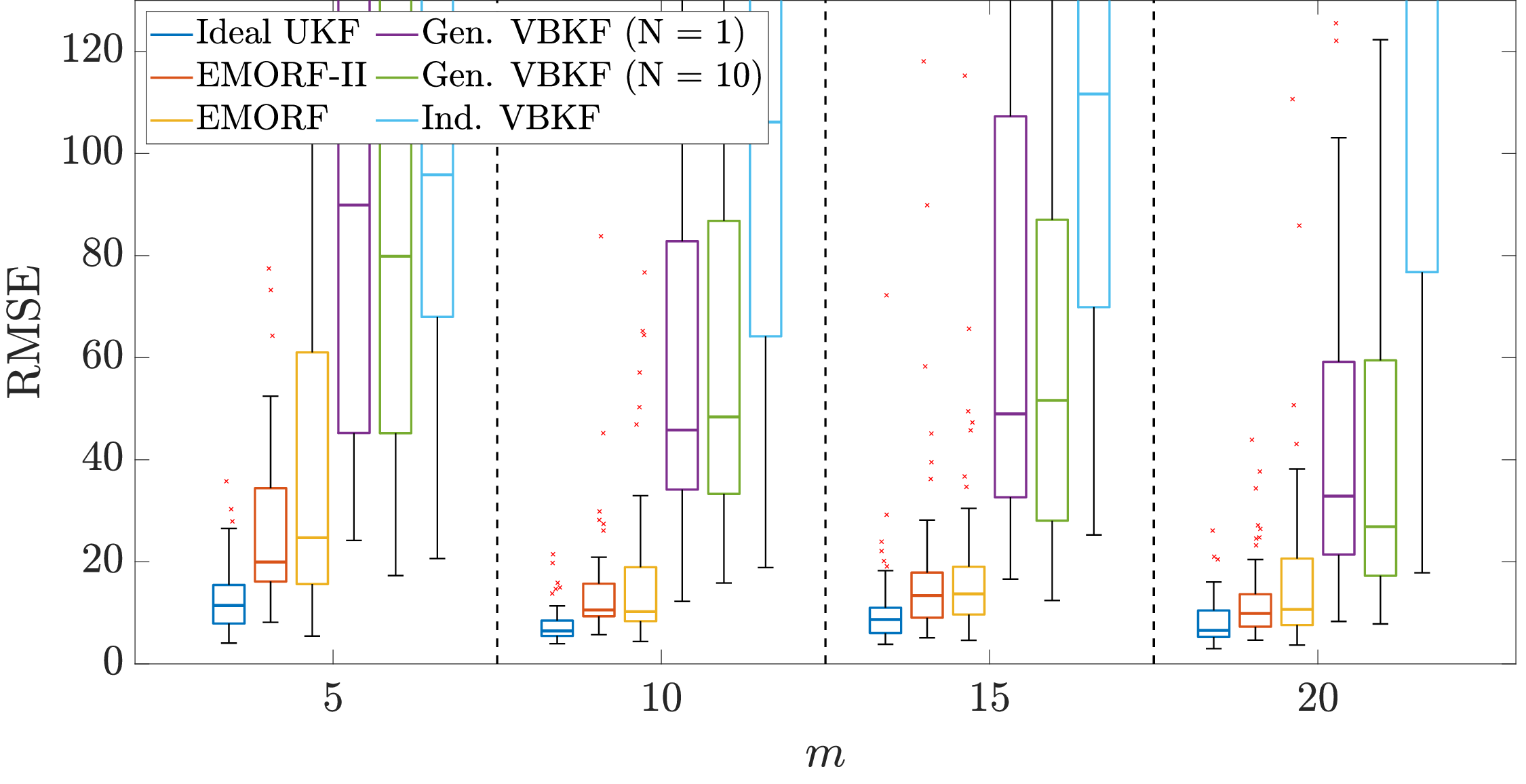} 
    \caption{RMSE comparison with $\lambda = 0.4 , \gamma = 1000$ and $m \in \{ 5, 10 ,15,20\}$.}
    \label{sim_2}
\end{figure}
In the next scenario, with $\lambda = 0.4$ and $\gamma = 1000$, we evaluate the error performance of each algorithm with increasing $m$. The resulting RMSE values are presented as box plots in Fig.~\ref{sim_2}. The results demonstrate that while all algorithms result in reduced RMSE as $m$ increases, EMORF-II consistently outperforms all the practical methods. This highlights the distinguishing feature of EMORF-II i.e. to adaptively learn outlier characteristics.   


\begin{figure}[h!]
    \centering
    \includegraphics[width=0.48\textwidth]{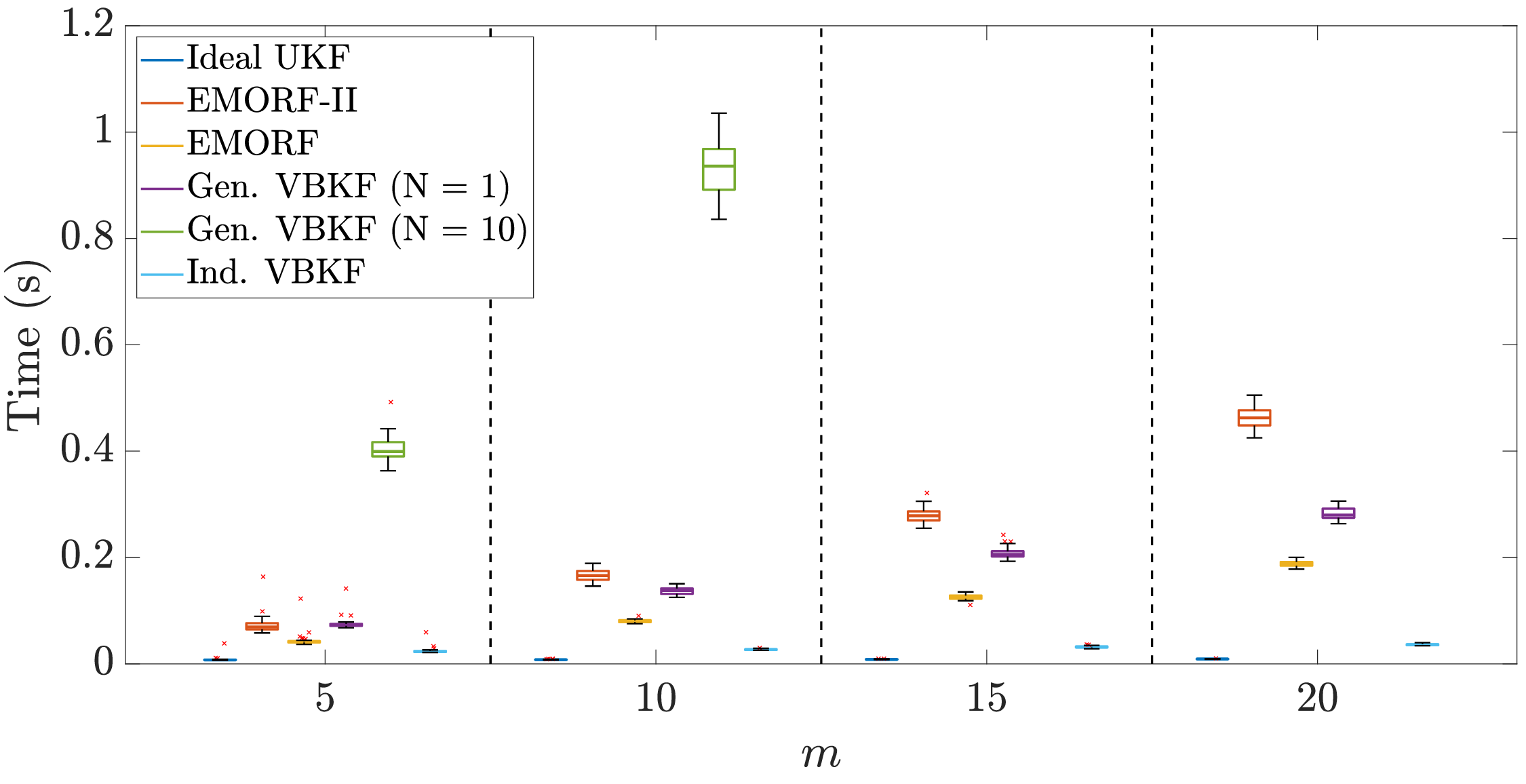} 
    \caption{Execution time comparison with $\lambda = 0.4, \gamma = 1000$ and $m \in \{ 5, 10 ,15, 20\}$.}
    \label{sim_3}
\end{figure}
In the third setup, we evaluate the computational efficiency of each algorithm by measuring their running times with increasing $m$, keeping the other simulation parameters identical to the previous experiment. The distributions of execution times across all methods are presented in Fig.~\ref{sim_3}. The results indicate that EMORF-II exhibits the highest computational cost, second only to Gen. VBKF ($N = 10$). The increased overhead, as compared to EMORF, is primarily due to the additional processing required to estimate $\hat{b}_k$, which facilitates adaptive learning of outlier statistics. Nevertheless, the computational complexity remains in the same order: \( \mathcal{O}(m^4) \) as in EMORF and Gen. VBKFs. This can be attributed to computing determinants and inverses of \( m \times m \) matrices with complexity \( \mathcal{O}(m^3) \) for each \( \mathcal{I}_k^i \) (for \( i \in \{1, \ldots, m\} \)).


\section{Conclusion}
We presented an outlier-robust filter, namely EMORF-II, for a general case where the measurement noise is correlated. Using insights from ASOR, we modify the structure of EMORF to devise EMORF-II enabling it to learn outlier characteristics during inference along with outlier detection. Numerical experiments under different scenarios verify that EMORF-II is superior in terms of error performance compared with similar state-of-the-art methods. However, the improved performance comes at a price of more computational overhead. Nevertheless, the complexity order remains $\mathcal{O}(m^4)$ as exhibited by other algorithms considering correlated measurement noise. This makes EMORF-II a useful candidate for a range of filtering applications.


\label{sec:ref}

\bibliographystyle{IEEEbib}
\bibliography{strings,refs}
\vfill
\newpage

\end{document}